\documentclass[conference]{IEEEtran}
\IEEEoverridecommandlockouts

\usepackage{cite}
\usepackage{amsmath,amssymb,amsfonts}
\usepackage{algorithmic}
\usepackage{graphicx}
\usepackage{textcomp}
\usepackage{xcolor}

\usepackage{booktabs}
\def\BibTeX{{\rm B\kern-.05em{\sc i\kern-.025em b}\kern-.08em
    T\kern-.1667em\lower.7ex\hbox{E}\kern-.125emX}}
\begin{document}

\title{WSD-MIL: Window Scale Decay Multiple Instance Learning for Whole Slide Image Classification\\

}

\author{\IEEEauthorblockN{Le Feng}
\IEEEauthorblockA{\textit{School of Artificial Intelligence} \\
\textit{Beijing University of Posts and Telecommunications}\\
Beijing, China \\
lefeng@bupt.edu.cn}
\and
\IEEEauthorblockN{Li Xiao}
\IEEEauthorblockA{\textit{School of Artificial Intelligence} \\
\textit{Beijing University of Posts and Telecommunications}\\
Beijing, China \\
andrewxiao@bupt.edu.cn
}
}

\maketitle

\begin{abstract}
In recent years, the integration of pre-trained foundational models with multiple instance learning (MIL) has improved diagnostic accuracy in computational pathology. However, existing MIL methods focus on optimizing feature extractors and aggregation strategies while overlooking the complex semantic relationships among instances within whole slide image (WSI). Although Transformer-based MIL approaches aiming to model instance dependencies, the quadratic computational complexity limits their scalability to large-scale WSIs. Moreover, due to the pronounced variations in tumor region scales across different WSIs, existing Transformer-based methods employing fixed-scale attention mechanisms face significant challenges in precisely capturing local instance correlations and  fail to account for the distance-based decay effect of patch relevance. To address these challenges, we propose window scale decay MIL (WSD-MIL), designed to enhance the capacity to model tumor regions of varying scales while improving computational efficiency. WSD-MIL comprises: 1) a window scale decay based attention module, which employs a cluster‑based sampling strategy to reduce computational costs while progressively decaying attention window-scale to capture local instance relationships at varying scales; and 2) a squeeze-and-excitation based region gate module, which dynamically adjusts window weights to enhance global information modeling. Experimental results demonstrate that WSD-MIL achieves state-of-the-art performance on the CAMELYON16 and TCGA-BRCA datasets while reducing 62\% of the computational memory. The code will be publicly available.
\end{abstract}


\section{Introduction}
In the field of cancer diagnosis and therapy, histopathological examination has traditionally depended on pathologists visually inspecting tissue sections with a light microscope. This manual approach is both time-consuming and labor-intensive, and its diagnostic accuracy is vulnerable to inter-observer variability arising from differences in experience and skill, potentially introducing bias~\cite{haggenmuller2025discordance,yang2025foundation}. Recent advances in digital pathology—particularly high-resolution slide scanners—allow microscopic tissue sections to be digitized as whole-slide images (WSIs), thereby furnishing large-scale datasets for deep-learning-driven automated pathology~\cite{song2023artificial,bera2019artificial}. Nevertheless, the ultra-high resolution of WSIs (often on the gigapixel scale), coupled with the scarcity of pixel-level annotations, makes it challenging to apply conventional supervised learning methods directly to WSI analysis and processing~\cite{lu2024visual,li2024generalizable}.

\begin{figure}[h]
	\includegraphics[width=\columnwidth]{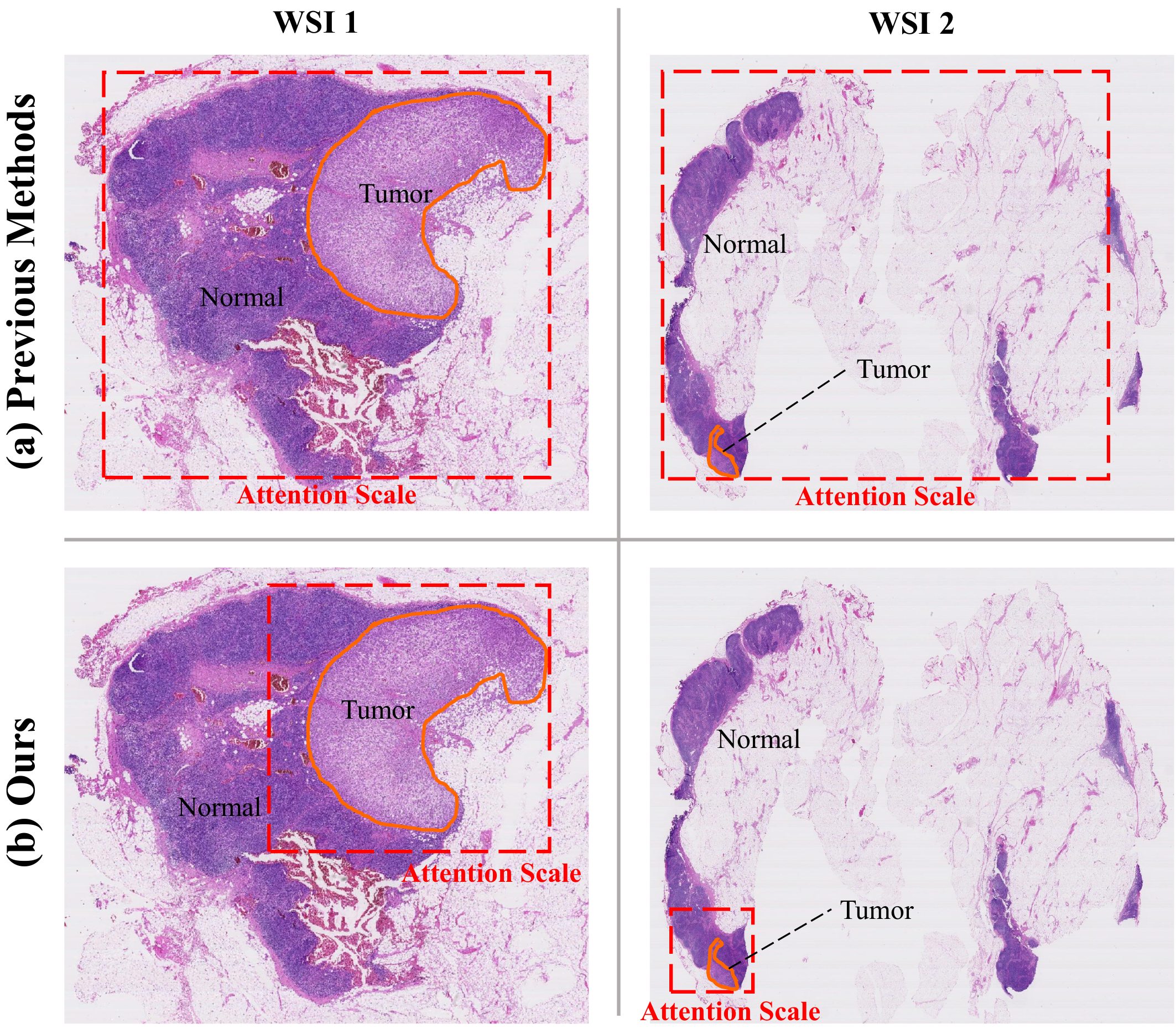}
	\caption{\textbf{Top:} The previous methods employed fixed-scale attention, which struggled to accurately capture key features in WSIs with significant scale variations in tumor regions. \textbf{Bottom:} The proposed method introduces a scale-decaying attention, enabling the model to flexibly adapt to tumor regions of different scales and thus precisely extract key features.} \label{fig-intro}
\end{figure}

To address these challenges, multiple-instance learning (MIL)—a weakly supervised framework that requires only slide-level labels—has been widely adopted for WSI diagnosis~\cite{li2023task,fillioux2023structured}. In MIL, each WSI is treated as a bag that is tiled into many non-overlapping patches serving as the bag’s instances~\cite{zhou2025robust,shi2024vila}. According to the standard MIL assumption, a bag is classified as negative if and only if all its instances are negative; otherwise, it is classified as positive. Traditional MIL methods generally involve two stages: firstly, all patches are embedded into feature vectors using a pre-trained encoder; secondly, these instance representations are aggregated into a bag-level representation for final classification by a classifier~\cite{liu2024semantics}. Existing research primarily focuses on two aspects: the pretraining of feature extractors and the design of MIL aggregator~\cite{wang2024rethinking, wang2023iteratively}. The former focuses on learning high-quality representations of pathological images. Foundation models, such as Virchow~\cite{vorontsov2024foundation} and Uni~\cite{chen2024towards}, leverage self-supervised algorithms like DINO~\cite{oquab2024dinov2} to extract morphological features. The latter, namely MIL aggregators, prioritizes the integration of patch-level features. Simple schemes such as mean pooling and max pooling are common, yet they perform poorly when positive patches are heavily outnumbered by negatives. Consequently, attention-based aggregators—including ABMIL~\cite{ilse2018attention}, CLAM~\cite{lu2021data}, and DTFD~\cite{zhang2022dtfd}—have been proposed to assign learnable weights to each instance. However, these methods typically embed each patch into the feature space independently and perform aggregation only at later stages, failing to capture spatial relationships between patches. This limitation restricts the model’s ability to comprehensively represent the overall pathological structure~\cite{yang2024mambamil}.

In recent years, Transformer-based methods have attracted considerable attention~\cite{shi2025positional}. These approaches employ multi-layer self-attention mechanisms between the frozen feature extractor and the aggregator, facilitating the effective modeling of nonlinear dependencies among instances~\cite{shao2021transmil}. However, given that each WSI comprises tens of thousands of patches, Transformer-based methods incur substantial computational and memory overhead, typically exhibiting $O(n^2)$ complexity. This renders them impractical for large-scale WSI analysis~\cite{han2022survey,chu2024retmil}. Additionally, recent studies have shown that task-relevant patch features comprise only a minute fraction of the total patch set; applying global self-attention indiscriminately to all instances can homogenize their representations, thereby diluting the contribution of these critical local features~\cite{WANG2023102645}. To address these challenges, the R2T has explored region-based instance dependency modeling, which captures feature correlations within local regions~\cite{tang2024feature}. Nevertheless, this approach predominantly relies on fixed-scale region attention, which fails to account for tumor regions’ scale variations across different WSIs, thereby limiting flexibility and accuracy in characterizing diverse tumor features, as illustrated in Fig.~\ref{fig-intro}(a). Furthermore, cancer cells tend to exhibit a clustered distribution, suggesting that patch-to-patch correlations should decay with increasing spatial distance. However, conventional fixed-scale attention mechanisms treat all patches within a region equally, failing to effectively model the varying significance of local patches. Therefore, it is crucial to develop a modeling strategy that can adapt to varying tumor scales, dynamically adjust correlations, as illustrated in Fig.~\ref{fig-intro}(b), and improve computational efficiency, ultimately enhancing both the accuracy and efficiency of WSI processing.

In this work, we propose an innovative window scale decay multiple instance learning (WSD-MIL) framework for WSI classification. This framework integrates two key components: a window scale decay based attention module (WSDA) and a squeeze-and-excitation based region gate module (SERG). The WSDA first applies a clustering algorithm to analyze the instance embedding vectors, leveraging a feature sampling strategy within each cluster to reduce computational cost while preserving feature diversity and representativeness. Subsequently, the module employs a progressively decaying window-scale attention mechanism to model local regions, capturing the correlations between patches in tumor regions of varying scales within WSI. Moreover, the module dynamically adjusts attention intensity based on the relative distances between instances. To further enhance the capacity for global information modeling, we introduce a SERG module, which captures dependencies among different regional partitions on a global scale and dynamically assigns varying weights to different regions. This design facilitates a comprehensive feature optimization process that transitions progressively from a global scale to a local scale and back to the global scale. The contributions of this work are summarized as follows:
\begin{itemize}
	\item A novel window scale decay based attention module is proposed, utilizes a feature cluster-based sampling strategy to reduce computational overhead, while employing a progressively decaying window-scale attention mechanism to model local regions, effectively capturing the correlations between patches in tumor regions of varying scales within WSI.
	\item A new squeeze-and-excitation based region gate module is proposed, which assigns different weights to different regional blocks to capture the dependencies among them on a global scale, thereby further enhancing the capacity for global information modeling.
	\item We conducted extensive experiments on two datasets, including Camelyon16 and TCGA-BRCA. The results demonstrated the effectiveness of the proposed method.
\end{itemize}

\section{Method}

\subsection{Problem formulation}

\begin{figure*}[h]
	\includegraphics[width=\textwidth]{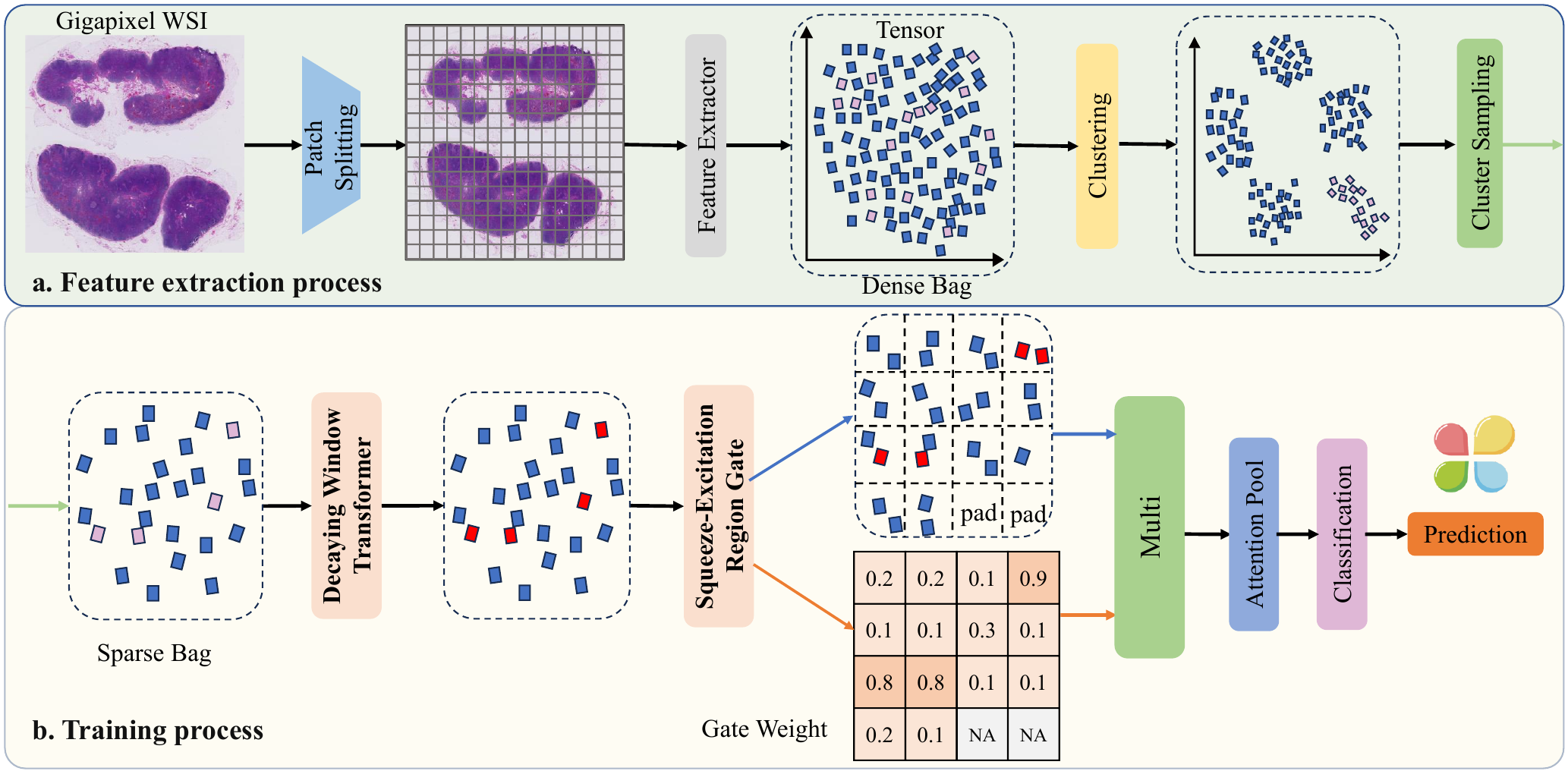}
	\caption{Overview of the proposed WSD-MIL.} \label{fig1}
\end{figure*}

\begin{figure*}[h]
	\includegraphics[width=\textwidth]{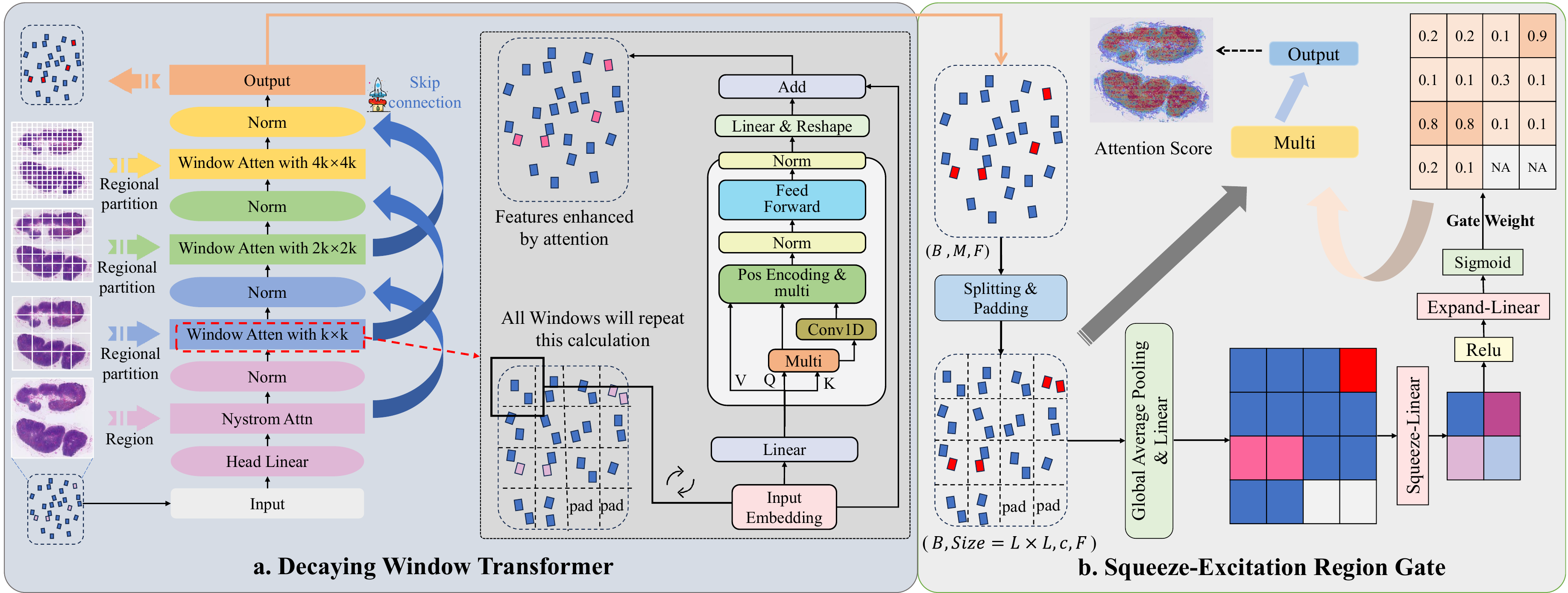}
	\caption{The structures of the decaying window transformer and the squeeze-excitation region gate.} \label{fig2}
\end{figure*}

Given a dataset of $N$ whole-slide images (WSIs) $D = \left\{ {\left( {{X_i},{y_i}} \right)} \right\}_{i = 1}^N$, where the slide-level label ${y_i} \in \left\{ {0, \cdots ,C - 1} \right\}$, our objective is to learn a mapping:
\begin{equation}
	f:X_i \mapsto \hat y_i , {\hat y_i} = f({X_i}) \approx {y_i}
\end{equation}
Under the multiple-instance learning (MIL) paradigm, each WSI is subdivided into $n$ non-overlapping patches, yielding a bag ${B} = \left\{ {{x_{1}}, \cdots ,{x_{{n}}}} \right\}$. Following the standard MIL assumption, a bag is positive if at least one patch is cancerous, and negative otherwise~\cite{shao2024multi}. Every patch is encoded by a pretrained network $E\left(  \cdot  \right)$ to obtain instance embeddings ${z_{j}} = E\left( {{x_{j}}} \right) \in {\mathbb{R}^d}$, aggregated as ${\mathcal{Z}} = \left\{ {{z_{1}}, \cdots ,{z_{{n}}}} \right\}$. A permutation-invariant aggregator $A\left(  \cdot  \right)$ produces a bag-level representation ${h} = A\left( {{\mathcal{Z}}} \right)$, which is then passed to a classifier $g\left(  \cdot  \right)$ to yield the prediction
\begin{equation}
	\begin{aligned}
		{\hat y} = g\left( {{h}} \right) = g\left( {A\left( {E\left( {{X}} \right)} \right)} \right)
	\end{aligned}
\end{equation}
However, conventional MIL approaches typically (i) overlook the complex spatial and semantic dependencies among patches, relying solely on simple pooling or attention-based aggregation and therefore failing to capture the macroscopic tissue architecture; and (ii) incur prohibitive computational overhead when processing tens of thousands of redundant patches, causing critical information to be drowned out. In the remainder of this paper, we introduce a framework that simultaneously models long-range dependencies and dynamically selects salient regions to alleviate these two bottlenecks.

\subsection{Overall architecture of window scale decay multiple instance learning framework (WSD-MIL)}

The WSD-MIL proposed in this study is illustrated in Fig.~\ref{fig1}. First, during the feature extraction stage (Fig.~\ref{fig1}a), the WSI is partitioned into non-overlapping small patches at a 20× resolution, and instance features are extracted using a pretrained feature extractor. Next, in the model training stage (Fig.~\ref{fig1}b), the extracted features are fed into the window-scale decay based attention module (WSDA). This module first employs a cluster‑based sampling strategy to eliminate redundant patch-level features and then utilizes a decaying window Transformer to precisely model the correlations among instances within tumor regions at different scales. Subsequently, all regional features are processed by the squeeze-and-excitation based region gate module (SERG), which dynamically assigns weights to different window regions to capture global inter-regional dependencies. Finally, a WSI-level attention aggregator and a linear classifier are used to predict bag-level labels.

\subsection{Window scale decay based attention module (WSDA)}\label{AA}
The WSDA module is designed to enhance the model’s ability to capture inter-instance relationships within tumor regions of varying scales in WSIs while reducing computational cost. This module consists of two key components: feature clustering and sampling, and decaying window transformer.

\paragraph{Feature Clustering-based Sampling Strategy} High-resolution WSI typically yields tens of thousands of largely redundant patches, creating substantial computational overhead. To curb this burden, each WSI is first tiled at 20$\times$ magnification into non-overlapping patches and fed to a pretrained encoder (e.g., ResNet-50 or Virchow). This produces an instance-level feature set 
\begin{equation}
	\begin{aligned}
		\mathcal{Z} = \left\{ {{z_i}} \right\}_{i = 1}^n,{z_i} \in {\mathbb{R}^d}
	\end{aligned}
\end{equation}
where $z_i$ is the $i$-th patch embedding and $n$ is the total number of patches extracted from the slide. To partition these embeddings, we apply $K$-means clustering~\cite{hamerly2003learning} by minimizing
\begin{equation}
	\begin{aligned}
		J = \sum\limits_{k = 1}^\mathcal{K} {\sum\limits_{{z_j} \in {C_k}}^{} {{{\left\| {{z_j} - {\mu _k}} \right\|}^2}} }
	\end{aligned}
\end{equation}
where $C_k$ denotes the $k$-th cluster and $\mu_k$ is centroid. Each cluster $C_k = \{ z_{k,1}, \dots, z_{k,n_k} \}$ contains $n_k$ feature vectors, with $\sum\limits_{k = 1}^{\cal K} {{n_k}}  = n$. 

To further reduce complexity while preserving tissue heterogeneity, we perform stratified random sampling: a fixed proportion $\alpha $\% of features is kept from every cluster, yielding subsets ${C'_k} \subset {C_k}$. The final, compact feature pool for downstream analysis is then
\begin{equation}
	Z = \mathop  \cup \limits_{k = 1}^{\cal K} {C'_k}
\end{equation}
This cluster-aware sampling strategy lowers memory and runtime demands yet retains a balanced representation of the slide’s diverse histologic patterns, thereby improving overall computational efficiency without compromising feature diversity.


\paragraph{Decaying Window Transformer} The overall architecture of the decaying window Transformer is illustrated in Fig.~\ref{fig2}a. For an input feature matrix $Z\in\mathbb{R}^{B\times M \times F}$, where $B$ represents the batch size, $M$ denotes the number of instances, and $F$ is the feature dimension, a linear projection is first applied to obtain the query, key, and value matrices: 
\begin{equation}
	[Q, K, V] = ZW_{qkv}, W_{qkv} \in \mathbb{R}^{F \times 3F}
\end{equation}
Next, regional sampling is performed on $Q$ and $K$ to obtain sampled queries and keys, denoted as $Q_m$ and $K_m$, respectively. To reduce computational complexity, a Nyström layer~\cite{xiong2021nystromformer} is employed to approximate global attention. The Moore-Penrose pseudoinverse, denoted by $(·)^+$, is used in the computation of Nyström attention, which is formulated as follows to obtain the attention matrix $H$:
\begin{equation}
\begin{aligned}
	H= Linear\Big(softmax(QK_m^{\rm{T}}) ({softmax(Q_mK_m^{\rm{T}})})^{+}\\ softmax(Q_mK^{\rm{T}}) V\Big)
\end{aligned}
\end{equation}
Next, we perform multi-head attention operations at window scales of $k \times k$, $2k \times 2k$, and $4k \times 4k$ ($k = 4$). For the $4 \times 4$ window attention process, as shown in the right half of Fig.~\ref{fig2}a, the input feature matrix is transformed as follows:
\begin{equation}
	\begin{aligned}
		H \in \mathbb{R}^{B \times M \times F} \to H \in \mathbb{R}^{B \times 4 \times 4 \times c \times F}
	\end{aligned}
\end{equation}
where $c$ represents the number of instances in each window, and it satisfies $4 \cdot 4 \cdot c = M$. The feature of each individual window is denoted as $H_l \in \mathbb{R}^{B \times c \times F}$, where $l \in {1, 2, \cdots, 16}$.
This partitioning strategy ensures that instance features within each local window maintain spatial adjacency, facilitating the extraction of fine-grained regional semantic information. Within each window region $H_l$, standard multi-head self-attention is applied to enhance the modeling ability of correlations between instances within the region: 
\begin{equation}
\begin{aligned}
	{A_l} = softmax\left( \frac{{Q_l K_l^{\rm{T}}}}{{\sqrt {d_k}}} + P \right) V_l,\\ P = Conv(Q_l K_l^{\rm{T}}), l=1, \cdots ,16
\end{aligned}
\end{equation}
where $Q_l$, $K_l$, and $V_l$ are obtained by linear transformations of $H_l$, $d_k$ is the scaling factor, and $P$ is the positional encoding information obtained through a 1D convolution. The resulting output is normalized and linearly transformed, followed by a residual connection with the features from the previous layer:
\begin{equation}
	{H'} = H + Linear\left( Norm\left( Linear\left( Norm\left( A \right) \right) \right) \right)
\end{equation}
The window scale is then gradually reduced to capture finer-grained features. The entire window-decaying Transformer process can be summarized as: 
\begin{equation}
	\tilde H = R_{4k}(R_{2k}(R_k(Nys(H \in \mathbb{R}^{B \times M \times F}))))
\end{equation}
where $R_{4k}$, $R_{2k}$, and $R_k$ represent the window attention mechanisms at scales of $4k \times 4k$, $2k \times 2k$, and $k \times k$, respectively, and $Nys$ refers to the Nyström attention mechanism.

\subsection{Squeeze-and-excitation based region gate module (SERG).}
Although the WSDA module captures correlations between patches from coarse to fine granularity, different regions contribute unequally to classification outcomes. In particular, regions containing positive instances exhibit highly distinctive features, whereas negative regions often provide little informative value. Thus, a mechanism is required to emphasize these discriminative regions while suppressing uninformative ones. The SERG module is designed to assign adaptive weights to different regions, capturing inter-region dependencies at a global scale and further enhancing global feature modeling capabilities. The overall architecture of the SERG is illustrated in Fig.~\ref{fig2}b.

Given the input feature matrix $H'\in {\mathbb{R}^{B\times M \times F}}$, it is partitioned into $L\times L$ windows $\{\tilde H_l\}_{l=1}^{L^2}$, where we set $L = 8$. First, global average pooling $GAP$ is applied to each window $\tilde H_l$:
\begin{equation}
	{Z'_l} = GAP\left( {{{\tilde H}_l}} \right),l=1,\cdots,L^2
\end{equation}
To assign different weights to different regions, we adopt the Squeeze-and-Excitation (SE) mechanism~\cite{hu2018squeeze}, which consists of squeeze and excitation steps. For squeeze step, pooled feature vectors are projected into a low-dimensional embedding space via a linear transformation:
\begin{equation}
	e = \sigma \left( {{W_1}{{Z'}}} \right),{W_1} \in \mathbb{R}^{L^2//r\times L^2}
\end{equation}
where $Z' =  \cup _{l = 1}^{L^2}{Z'_l}$, $r$ is the reduction ratio and $\sigma \left(  \cdot  \right)$ is the ReLU activation function.

For excitation step, the reduced representation is then expanded back to its original dimension:
\begin{equation}
	G = sigmoid\left( {{W_2}e} \right),{W_2} \in \mathbb{R}^{L^2\times L^2//r}
\end{equation}

Finally, the regional gating weights $G \in {\mathbb{R}^{L \times L}}$ are obtained and used for region-wise feature reweighting: ${H_{out}} = H' \odot G$, where $\odot$ denotes element-wise multiplication at the regional level. The weighted feature map $H_{out}$ is then propagated through subsequent network layers, enhancing inter-region modeling and improving feature representation. A WSI‑level attention aggregator followed and a linear classifier are ultimately employed to predict the bag‑level label from $H_{out}$.

\section{Experiments and Results}
\subsection{Datasets}
To verify the effectiveness of the proposed framework, we carried out experiments on two extensively cited public histopathology datasets: CAMELYON‑16~\cite{bejnordi2017diagnostic} and TCGA‑BRCA~\cite{petrick2021spie}. Both collections comprise whole‑slide images (WSIs) stained with haematoxylin and eosin (H\&E) and digitized using high‑resolution scanners. Owing to their broad adoption for developing and benchmarking cancer‑diagnosis algorithms, these datasets serve as authoritative reference standards in computational pathology. Detailed descriptions of the datasets and the experimental setup are provided in the following content.


\textit{1) CAMELYON-16:} The CAMELYON-16 dataset consists of 399 whole‑slide images (WSIs) of sentinel lymph‑node biopsies from breast‑cancer patients. Among these, 239 slides contain metastatic deposits (positive), whereas 160 do not (negative). The official split assigns 270 slides to training (162 positive, 108 negative) and 129 to testing (77 positive, 52 negative). Metastatic foci are delineated at the pixel level, and each slide also receives a slide‑level label from expert pathologists. In this study, we use only the slide‑level labels and adopt a weakly supervised learning paradigm that eschews pixel‑wise annotations. A key challenge of CAMELYON16 is that metastatic regions are often minute, rendering their localisation within the extensive benign tissue particularly difficult.

\textit{2) TCGA‑BRCA:} The TCGA‑BRCA cohort comprises 977 diagnostic WSIs of primary invasive breast carcinoma. We focus on the two predominant histological subtypes— invasive ductal carcinoma (IDC, 779 slides) and invasive lobular carcinoma (ILC, 198 slides). Slides are randomly divided on a per‑patient basis into a training set of 780 slides and a test set of 197 slides. Only slide‑level subtype labels extracted from pathology reports are available; no region‑level annotations are provided. Compared with CAMELYON16, tumour regions in TCGA‑BRCA usually occupy a larger proportion of the tissue section. However, pronounced morphological heterogeneity, abundant adipose stroma and inter‑scanner colour variation make subtype classification non‑trivial, positioning TCGA‑BRCA as a realistic yet challenging benchmark for weakly supervised breast‑cancer analysis.

\subsection{Experiment setup and evaluation metrics}
\begin{figure}[h]
	\includegraphics[width=\columnwidth]{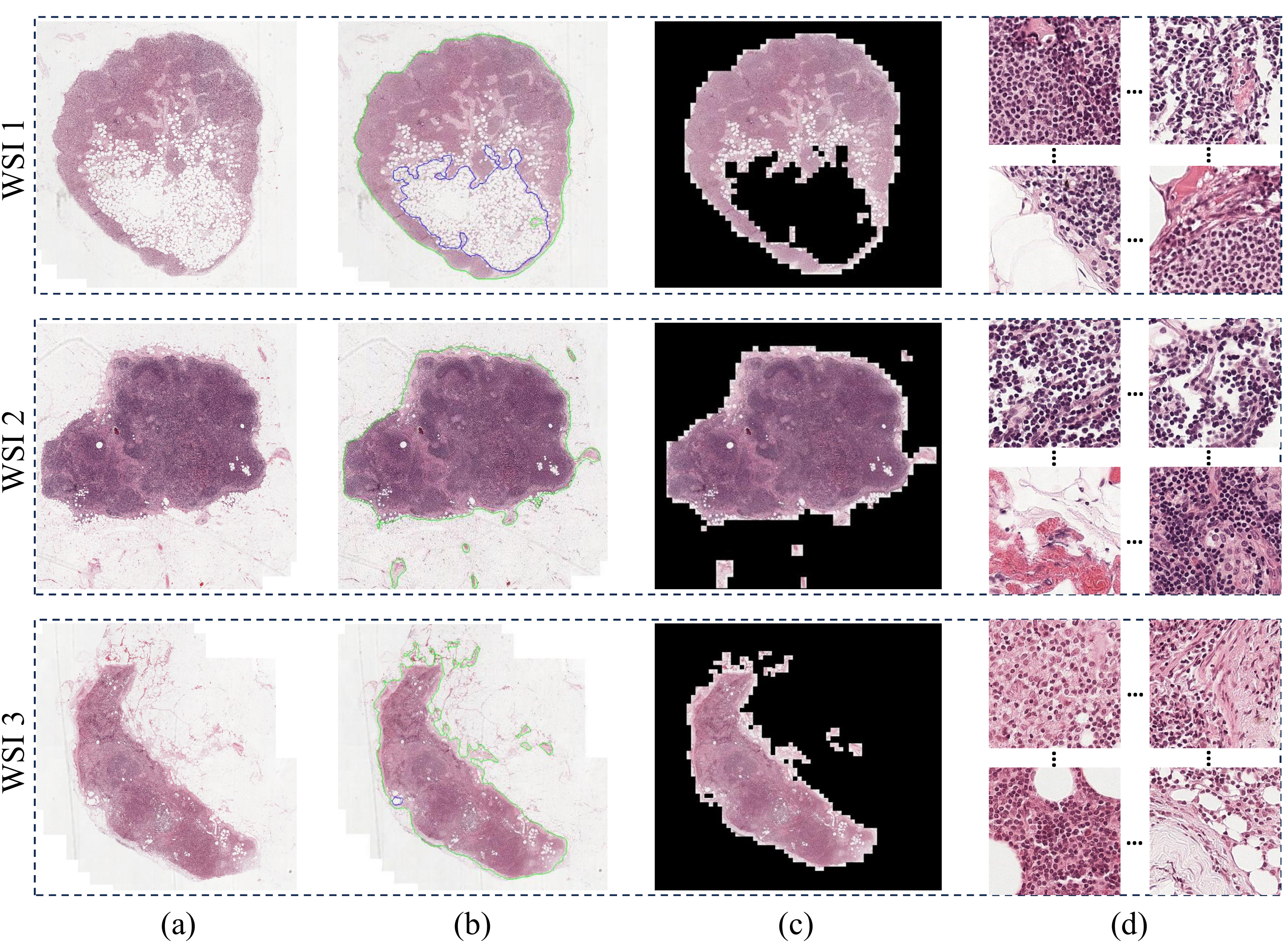}
	\caption{The image preprocessing steps for WSIs. (a) Original WSI samples; (b)The foreground region of the WSI obtained via threshold-based segmentation; (c) The WSI after background removal; (d) Multiple 256 × 256-pixel patches obtained by cropping the foreground of the WSI.} \label{fig3}
\end{figure}

\begin{table*}[h]
	\caption{The classification performance of the proposed method and the R2T-MIL method under patch sampling ratios of 100\%, 60\%, and 20\%, respectively.}
	\label{tab1}
	\centering
	\renewcommand{\arraystretch}{1} 
	\setlength{\tabcolsep}{12pt}
	\resizebox{\textwidth}{!}{
		\begin{tabular}{l l cccc}
			\toprule
			\textbf{Methods} & \textbf{Patch Sample $\alpha$\%} & \multicolumn{4}{c}{\textbf{Camelyon16 $_{(ResNet50)}$}} \\ 
			\cmidrule(lr){3-6} 
			& & Acc & AUC & F1 & Memory (GB)\\
			\midrule
			R2T-MIL~\cite{tang2024feature}+Sample & 100\% & 89.44$_{\pm2.80}$ & 91.38$_{\pm2.64}$ & 85.74$_{\pm3.98}$ & 12 (100\%) \\
			& 60\% & 89.70$_{\pm1.94}$ & \textbf{92.31$_{\pm1.89}$} & 86.03$_{\pm3.42}$ & 8 (66\%) \\
			& 20\% & \textbf{89.70$_{\pm1.35}$} & 92.04$_{\pm1.27}$ & \textbf{86.53$_{\pm2.58}$} & \textbf{5 (42\%)} \\
			\hline
			Ours+Sample& 100\% & 90.46$_{\pm1.93}$ & 93.59$_{\pm2.27}$ & 88.06$_{\pm1.69}$ & 13 (100\%) \\
			& 60\% & \textbf{91.21$_{\pm2.16}$} & \textbf{94.09$_{\pm2.66}$} & \textbf{88.96$_{\pm2.29}$} & 8 (61\%) \\
			& 20\% & 91.20$_{\pm3.25}$ & 92.74$_{\pm1.92}$ & 87.89$_{\pm3.69}$ & \textbf{5 (38\%)} \\
			
			\bottomrule
		\end{tabular}
	}
\end{table*}

\begin{table*}[h]  
	\caption{Comparison with state-of-the-art methods on the two datasets.}
	\label{tab2}
	\centering
	\renewcommand{\arraystretch}{1.5} 
	\resizebox{\textwidth}{!}{
		\begin{tabular}{l ccc ccc ccc}
			\toprule
			\textbf{Methods} & \multicolumn{3}{c}{\textbf{Camelyon16 $_{(ResNet50)}$}} & \multicolumn{3}{c}{\textbf{TCGA-BRCA $_{(ResNet50)}$}} & \multicolumn{3}{c}{\textbf{TCGA-BRCA $_{(Virchow)}$}} \\ 
			\cmidrule(lr){2-4} \cmidrule(lr){5-7} \cmidrule(lr){8-10}
			& Acc & AUC & F1 & Acc & AUC & F1 & Acc & AUC & F1 \\
			\midrule
			Mean Pooling & 87.18$_{\pm2.69}$ & 89.82$_{\pm2.56}$ & 82.28$_{\pm3.10}$ & 84.86$_{\pm3.84}$ & 89.43$_{\pm1.91}$ & 68.64$_{\pm7.09}$ & 93.04$_{\pm2.36}$ & 94.39$_{\pm3.29}$ & 82.99$_{\pm6.26}$ \\
			Max Pooling & 75.44$_{\pm2.83}$ & 74.07$_{\pm5.57}$ & 68.23$_{\pm5.15}$ & 82.19$_{\pm1.63}$ & 85.32$_{\pm4.52}$ & 64.39$_{\pm5.28}$ & 91.81$_{\pm1.51}$ & 91.85$_{\pm4.02}$ & 79.27$_{\pm5.45}$ \\
			ABMIL~\cite{ilse2018attention} & 87.94$_{\pm1.44}$ & 88.89$_{\pm2.92}$ & 81.78$_{\pm3.19}$ & 87.61$_{\pm2.28}$ & 88.89$_{\pm4.33}$ & 70.64$_{\pm5.30}$ & 91.91$_{\pm3.47}$ & 94.78$_{\pm3.15}$ & 82.32$_{\pm7.19}$ \\
			CLAM~\cite{lu2021data} & 89.58$_{\pm2.39}$ & 90.47$_{\pm2.78}$ & 84.68$_{\pm1.51}$ & 86.80$_{\pm2.74}$ & 89.29$_{\pm2.88}$ & 71.26$_{\pm5.74}$ & 93.05$_{\pm2.34}$ & 93.43$_{\pm3.23}$ & 82.63$_{\pm6.02}$ \\
			TransMIL~\cite{shao2021transmil} & 89.49$_{\pm2.71}$ & 90.67$_{\pm3.14}$ & 84.55$_{\pm3.14}$ & 89.77$_{\pm2.82}$ & 89.54$_{\pm3.90}$ & 75.58$_{\pm5.75}$ & 91.31$_{\pm3.75}$ & 93.88$_{\pm3.32}$ & 80.26$_{\pm8.54}$ \\
			S4-MIL~\cite{fillioux2023structured} & 88.69$_{\pm2.22}$ & 89.81$_{\pm1.39}$ & 84.66$_{\pm3.70}$ & 88.13$_{\pm3.46}$ & 90.56$_{\pm5.09}$ & 73.13$_{\pm8.39}$ & 91.81$_{\pm1.39}$ & 93.21$_{\pm3.91}$ & 80.49$_{\pm4.21}$ \\
			R2T-MIL~\cite{tang2024feature} & 89.70$_{\pm1.35}$ & 92.04$_{\pm1.27}$ & 86.53$_{\pm2.58}$ & 89.25$_{\pm2.17}$ & 89.44$_{\pm4.13}$ & 74.54$_{\pm4.41}$ & 92.51$_{\pm2.29}$ & 93.95$_{\pm3.12}$ & 82.60$_{\pm5.61}$ \\
			\textbf{Ours} & \textbf{91.20$_{\pm3.25}$} & \textbf{92.74$_{\pm1.92}$} & \textbf{87.89$_{\pm3.69}$} & \textbf{90.38$_{\pm1.64}$} & \textbf{91.14$_{\pm3.32}$} & \textbf{77.68$_{\pm3.96}$} & \textbf{93.35$_{\pm2.81}$} & \textbf{94.83$_{\pm3.15}$} & \textbf{84.19$_{\pm6.61}$} \\
			\bottomrule
		\end{tabular}
	}
\end{table*}

During preprocessing (see Fig.~\ref{fig3}), each whole‑slide image (WSI) is first subjected to threshold‑based segmentation to delineate the foreground region (Fig.~\ref{fig3}b); background pixels are then removed to generate a binary mask (Fig.~\ref{fig3}c). Only the foreground is subsequently tiled into $256 \times 256$‑pixel patches at $20\times$ optical magnification (Fig.~\ref{fig3}d). Each patch is finally embedded into a 1,024‑dimensional or 1,280‑dimensional feature vector by means of a pretrained ResNet‑50~\cite{he2016deep} and Virchow~\cite{vorontsov2024foundation} backbone, respectively. All experiments were conducted on one NVIDIA RTX 4090 GPU. The Adam optimizer was employed to update the model weights, with an initial learning rate set to 1e-5. The batch size was set to 1, with a total of 100 epochs. All other baseline methods followed the same experimental settings. We selected accuracy (Acc), area under the curve (AUC), and F1-score (F1) as evaluation metrics and reported their mean and standard deviation under five-fold cross-validation.

\subsection{Sample results}
To systematically evaluate the capacity of the cluster-based sampling strategy to curb resource consumption while preserving classification performance, we fixed the sampling ratio $\alpha$\% at 100\%, 60\%, and 20\% and conducted comparative experiments on two schemes: ``R2T‑MIL~\cite{tang2024feature} +Sample'' and ``the proposed WSD‑MIL (denoted as Ours+Sample)''. During the feature clustering process, the parameter $\mathcal{K}$ is set to 10. In every run, a frozen ResNet‑50 backbone served as the feature extractor, and all remaining training hyper‑parameters were held constant, ensuring that any performance differences arose solely from the number of sampled patches. Besides conventional metrics—Acc, AUC, and F1—we logged the GPU memory footprint (Memory, GB) after each training epoch to quantify the storage and computational burden introduced by the sampling strategy.

As reported in Table~\ref{tab1}, reducing the sampling ratio from 100\% to 20\% does not noticeably degrade the classification accuracy of either method, confirming that discarding large quantities of redundant patches has a negligible effect on discriminative power. At the 20\% setting, R2T‑MIL still attains 89.70\% Acc, 92.04\% AUC, and 86.53 F1, whereas WSD‑MIL reaches 91.20\% Acc, 92.74\% AUC, and 87.89 F1; moreover, WSD‑MIL achieves the table‑wide best AUC (94.09\%) at the 60\% ratio. Crucially, GPU memory usage decreases almost linearly with the sampling rate: at 20\% sampling, both methods require only 5 GB, representing savings of approximately 58\% (R2T‑MIL) and 62\% (WSD‑MIL) relative to their full‑sampling baselines. These findings demonstrate that the proposed clustering‑driven sampling mechanism effectively alleviates memory and computation bottlenecks while maintaining—or even slightly enhancing—classification metrics. Consequently, all subsequent experiments adopt the 20\% sampling ratio to balance efficiency and accuracy.

\subsection{Comparison with State-of-the-art methods}
\begin{table*}[h]
	\caption{Ablation study for WSD-MIL.}
	\label{tab3}
	\centering
	\renewcommand{\arraystretch}{1} 
	\setlength{\tabcolsep}{8pt}
	\resizebox{\textwidth}{!}{
		\begin{tabular}{l ccc ccc}
			\toprule
			\textbf{Methods} & \multicolumn{3}{c}{\textbf{Camelyon16 $_{(ResNet50)}$}} & \multicolumn{3}{c}{\textbf{TCGA-BRCA $_{(ResNet50)}$}} \\ 
			\cmidrule(lr){2-4} \cmidrule(lr){5-7}
			& Acc & AUC & F1 & Acc & AUC & F1 \\
			\midrule
			w/o WSDA & 88.18$_{\pm2.71}$ & 91.42$_{\pm2.04}$ & 83.78$_{\pm4.25}$ & 88.22$_{\pm1.37}$ & 89.10$_{\pm2.60}$ & 72.10$_{\pm3.43}$ \\
			w/ FixWin 8$\times$8 & 90.44$_{\pm2.80}$ & 92.37$_{\pm1.98}$ & 86.88$_{\pm4.37}$ & 89.87$_{\pm2.17}$ & 89.57$_{\pm2.83}$ & 74.65$_{\pm6.60}$ \\
			w/ FixWin 32$\times$32 & 89.93$_{\pm3.17}$ & 91.94$_{\pm2.19}$ & 87.10$_{\pm3.29}$ & 89.25$_{\pm3.26}$ & 89.18$_{\pm3.97}$ & 74.39$_{\pm8.46}$ \\
			w/o SERG & 90.46$_{\pm2.14}$ & 92.59$_{\pm1.61}$ & 87.69$_{\pm2.17}$ & 90.08$_{\pm2.76}$ & 90.83$_{\pm3.13}$ & 76.15$_{\pm7.11}$ \\
			w/o WSDA, SERG & 87.94$_{\pm1.44}$ & 88.89$_{\pm2.92}$ & 81.78$_{\pm3.19}$ & 87.61$_{\pm2.28}$ & 88.89$_{\pm4.33}$ & 70.64$_{\pm5.30}$ \\
			\textbf{Ours} & \textbf{91.20$_{\pm3.25}$} & \textbf{92.74$_{\pm1.92}$} & \textbf{87.89$_{\pm3.69}$} & \textbf{90.38$_{\pm1.64}$} & \textbf{91.14$_{\pm3.32}$} & \textbf{77.68$_{\pm3.96}$} \\
			\bottomrule
		\end{tabular}
	}
\end{table*}

To rigorously assess the applicability and superiority of WSD-MIL under diverse data distributions, feature extractors, and mainstream multiple-instance learning (MIL) frameworks, we conducted experiments on two representative public datasets: Camelyon16, which contains a low fraction of metastatic tissue, and TCGA-BRCA, noted for its pronounced morphological heterogeneity. Frozen features were extracted using a shared ResNet-50 backbone for both datasets and, for TCGA-BRCA only, the state-of-the-art pathology foundation model Virchow. Keeping all experimental conditions identical—training hyper-parameters, five-fold cross-validation, and evaluation metrics (Acc, AUC, F1)—we compared the proposed method against seven representative baselines: Simple aggregation (Mean Pooling, Max Pooling), Attention-based MIL (ABMIL, CLAM, S4-MIL), Fixed-scale transformer MIL (TransMIL, R2T‑MIL). This hierarchy—from conventional pooling, through fine-grained attention and fixed-scale transformers, to our adaptive-decay transformer—allows us to objectively quantify the gains of WSD-MIL in modeling local-to-global correlations and coping with tumor‑scale variability.

Table~\ref{tab2} summarizes the comparative results and shows that WSD-MIL delivers state-of-the‑art performance across all experimental settings. On Camelyon16 with a ResNet-50 backbone, it attains 91.20\% accuracy, 87.89\% F1, and 92.74\% AUC—improvements of 1.5\% and 1.4\% over R2T-MIL in accuracy and F1, and $ \approx $ 2\% over TransMIL in AUC—highlighting its heightened sensitivity to sparse metastatic foci. When transferred to the morphologically heterogeneous TCGA‑BRCA cohort using the same ResNet‑50 features, WSD‑MIL maintains a clear advantage (90.38\% accuracy, 91.14\% AUC, 77.68\% F1), exceeding the conventional attention model ABMIL by $ > $7\% in F1 and confirming its robustness in complex tumour contexts. Substituting the ResNet-50 features with high-dimensional semantic embeddings from the Virchow foundation model further raises overall scores, yet WSD-MIL still leads (93.35\% accuracy, 94.83\% AUC, 84.19\% F1), underscoring its strong synergy with large‑scale pretrained representations. Collectively, these findings demonstrate that the proposed adaptive mechanism—progressive attention-window decay coupled with region‑gated weighting—effectively overcomes the scale inflexibility of fixed‑scale transformers and sets new benchmarks for weakly supervised whole‑slide image classification.

\subsection{Ablation study}
To quantitatively evaluate the contributions of the two key components in WSD‑MIL—Window Scale Decay Attention (WSDA) and the Squeeze‑and‑Excitation Region Gate (SERG)—we designed five ablation configurations on the Camelyon16 and TCGA‑BRCA datasets: (i) removing WSDA (w/o WSDA); (ii) replacing WSDA with fixed 8 × 8 window self‑attention (FixWin 8×8); (iii) replacing WSDA with fixed 32 × 32 window self‑attention (FixWin 32×32); (iv) removing SERG (w/o SERG); and (v) removing both WSDA and SERG (w/o WSDA, SERG). Except for these alterations, all other experimental settings matched those of the complete model: a frozen ResNet‑50 for feature extraction, five‑fold cross‑validation, and evaluation using accuracy (Acc), area under the ROC curve (AUC), and F1‑score.

Table~\ref{tab3} shows that both modules markedly boost performance. Relative to the complete model, removing WSDA on Camelyon16 lowers Acc/AUC/F1 by 3.0/1.3/4.1 percentage points, respectively. Replacing WSDA with fixed 8 × 8 or 32 × 32 window self‑attention is slightly better than full removal, yet still trails the baseline by 0.8–1.3\% in accuracy and 0.6–1.0\% in F1, indicating that fixed scales cannot adequately model multi‑scale tumour associations. Eliminating SERG primarily affects global discriminative power: F1 drops from 87.89\% to 87.69\%, while Acc declines by 0.7\%. Ablating both WSDA and SERG further degrades performance to 87.94\% Acc and 81.78\% F1, confirming their synergistic benefit. A similar trend emerges on the more morphologically heterogeneous TCGA‑BRCA dataset: removing WSDA causes a 5.6\% F1 decline, removing SERG costs 1.5\%, and removing both results in a 7.0\% F1 reduction. In summary, WSDA is critical for capturing local correlations across scales, whereas SERG enhances class separability via global region re‑weighting; in concert, they deliver consistent gains of roughly 3\% in accuracy and 5–7\% in F1 across both datasets.

\section{Conclusion}
In this paper, we propose a window scale decay multiple instance learning framework, termed WSD-MIL. This framework utilizes a cluster-based sampling strategy to reduce computational overhead, while employing a progressively decaying window-scale attention mechanism to model local regions, effectively capturing the correlations between patches in tumor regions of varying scales within WSIs. Additionally, we design a squeeze-and-excitation based gate module, which dynamically adjusts the weights of different window regions to capture the dependencies between local regions across the global scope, thereby enhancing the model's ability to capture global information. Finally, we compare our approach with Transformer-based methods, and the experimental results demonstrate that WSD-MIL achieves state-of-the-art performance while significantly reducing computational costs.

\bibliographystyle{IEEEtran}         
\bibliography{IEEEabrv,reference}

\end{document}